\documentclass{article}
\usepackage{spconf,amsmath,graphicx}
\usepackage[table,xcdraw]{xcolor}
\usepackage{placeins}
\usepackage{hyperref}       

\newcommand{\comment}[1]{}

\title{Unsupervised Multimodal Language Representations using Convolutional Autoencoders}
\name{\href{https://orcid.org/0000-0002-4466-463X}{Panagiotis Koromilas\hspace{1mm}\includegraphics[scale=0.06]{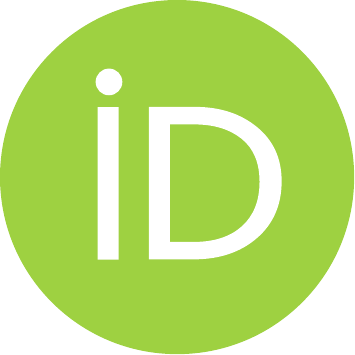}} \qquad Theodoros Giannakopoulos}
\address{Institute of Informatics \& Telecommunications, NCSR - Demokritos, Greece}
\begin{document}
\ninept
\maketitle
%
\begin{abstract}
Multimodal Language Analysis is a demanding area of research, since it is associated with two requirements: combining different modalities and capturing temporal information. During the last years, several works have been proposed in the area, mostly centered around supervised learning in downstream tasks. In this paper we propose extracting unsupervised Multimodal Language representations that are universal and can be applied to different tasks. Towards this end, we map the word-level aligned multimodal sequences to 2-D matrices and then use Convolutional Autoencoders to learn embeddings by combining multiple datasets. Extensive experimentation on Sentiment Analysis (MOSEI) and Emotion Recognition (IEMOCAP) indicate that the learned representations can achieve near-state-of-the-art performance with just the use of a Logistic Regression algorithm for downstream classification. It is also shown that our method is extremely lightweight and can be easily generalized to other tasks and unseen data with small performance drop and almost the same number of parameters. The proposed multimodal representation models are open-sourced and will help grow the applicability of Multimodal Language. 
\end{abstract}


\section{Introduction}
Multimodal Language Analysis focuses on extracting information from temporal language sequences using acoustic, visual and textual information. The two core application-specific problems of this area are Multimodal Sentiment Analysis, where the goal is to determine whether a utterance contains positive or negative sentiment, and Multimodal Emotion Recognition in which the objective is to predict the underlying emotion of the utterance. In the past years, research has been conducted in these problems, resulting in powerful models that achieve great performance on the particular tasks. Different types of fusion approaches, such as early fusion \cite{nojavanasghari2016deep}, memory fusion \cite{zadeh2018memory, zadeh2018multimodal}, multistage fusion \cite{liang2018multimodal} and tensor fusion \cite{zadeh2017tensor} have been examined. Modifications on LSTM architectures for multi-view learning \cite{rajagopalan2016extending} or context-dependent analysis \cite{poria2017context} have also been proposed, as well as the concept of attention on recurrent networks \cite{zadeh2018multi}, context-aware attention \cite{chauhan-etal-2019-context} and some transformer architectures \cite{tsai2019multimodal} have all been researched in depth.
 
However, most of the works treat the learning process in a supervised fashion. The individual subdomains (i.e. the different applications and datasets) of Multimodal Language have the particularity of a high level of variability in the recording conditions and the experimental setup in general. For example the most widely used dataset for Emotion Recognition, IEMOCAP \cite{busso2008iemocap}, is recorded in a laboratory with organized camera networks, while MOSEI \cite{zadeh2018multimodal}, the largest dataset for Multimodal Sentiment Analysis, is collected from Youtube videos. Thus, strongly supervised methods are of limited use and cannot generalize to unseen recording set-ups and different tasks. 

One of the few works that includes an unsupervised factor in their hybrid (including supervised and unsupervised factors) loss function is presented in \cite{pham2019found} were the trained representations are sequential which limits the usage of such embeddings to sequential architectures. Another method that learns unsupervised representations is introduced in \cite{sun2020learning}, where the proposed architecture extracts text-based embeddings and thus the information of other modalities is only used to enrich the textual information. However, the resulted representations of \cite{pham2019found} and \cite{sun2020learning} cannot properly model tri-modal interactions, since there is always a modality that interacts with the bi-modal representation of the remaining modalities rather than the actual input sequences. Adding that such approaches make the overall architecture task or data (eg. \cite{delbrouck2020transformer}) specific and that they do not test the generalization ability of the produced representations to other tasks and data set-ups, we have no proof for their performance on unseen data and tasks.

In this work, we propose a simple, yet powerful (in terms of both performance and computational efficiency), unsupervised  Multimodal Language representation scheme that adopts a convolutional autoencoder architecture to discover multimodal relationships between aligned representations of audio, text and visual aligned feature sequences. The core contributions of the proposed method are the following: 
\begin{enumerate}
    \item To our knowledge, this is the first method in the field of general Multimodal Language Analysis that is both multimodal in all three modalities and unsupervised 
    \item The proposed architecture is extremely transferable to other domains without negative impact neither on the performance or in the number of model parameters used. External experimentation proves that the performance is just slightly reduced when transferring knowledge from one dataset to another.  And this happens without retraining the representation method itself, just using its embeddings in the target domain and  classified by a simple logistic regression classifier. 
    \item The performance is competitive related to the SotA, with a major advantage with regards to the complexity of the proposed method: the proposed multimodal representation model has around 200K parameters. 
    \item The proposed method can even be used, with similar advantages (ultralight and accurate), to extract compact and low-dimensional unimodal or bimodal representations from a temporal sequence. 
    \item This representation is openly available through an easy-to-use model (\href{https://github.com/lobracost/mlr}{https://github.com/lobracost/mlr}). The ultra-light pretrained Multimodal Language Analysis model (1MB of size) can be directly applied without retraining on similar aligned datasets. 
\end{enumerate}

\begin{figure*}[h]
\includegraphics[width=.8\paperwidth,height=3cm]{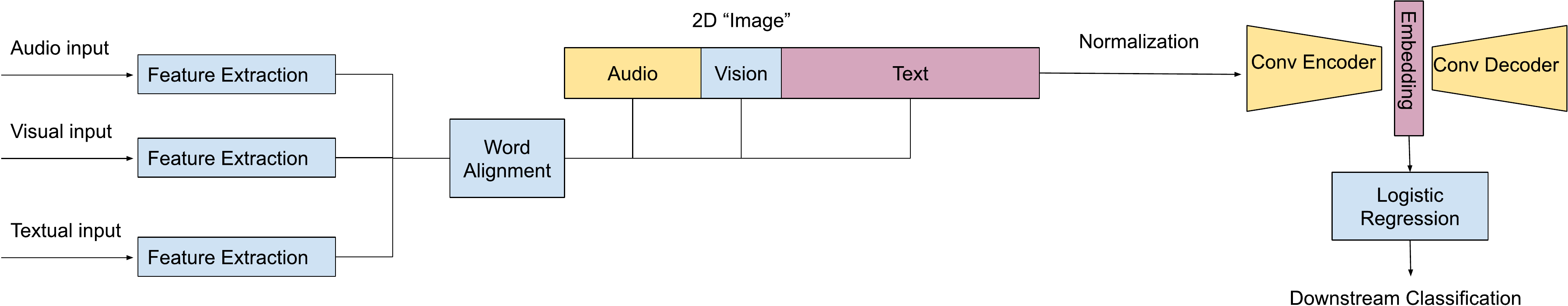}
    \caption{Word-level alignment is performed among the feature sequence of each modality resulting in a 2-D tensor (``image'') for each utterance. After proper normalization, a Convolutional Autoencoder is fitted in order to learn unsupervised representations. A Logistic Regression algorithm then trains a model on the downstream task using the produced embeddings as features. }
    \label{fig:architecture}
\end{figure*}

\section{Convolutional Autoencoder for multimodal sequences}

\subsection{2-D multimodal sequence representation}

In this paper we follow a widely used approach for basic unimodal feature extraction and multimodal alignment, similar to the one in various proposed methods, such as \cite{zadeh2018multi, zadeh2018memory, zadeh2017tensor, wang2019words, zadeh2018multimodal, liu2018efficient, tsai2019multimodal, tsai2019learning,  chauhan-etal-2019-context, sun2020learning}. More specifically, after extracting the features on each modality(visual, textual and aural), the procedure of word-level alignment that was firstly used for this task in \cite{tsai2019learning}, is performed. That is, the aligned video and audio
features are obtained by computing the expectation of their modality feature values over each word utterance
time interval.

The feature extraction procedure results in a time sequence of feature vectors for each modality that can be formalized by a $N$ x $M$ matrix, where $N$ is the number of timestamps and $M$ is the number of features. Due to the performed alignment, sequences of all modalities include the same number of timestamps $N$ and thus they can be combined in an $N$ x $(M_{audio} + M_{vision} + M_{text})$ multimodal matrix $X$. This merged matrix is the selected initial multimodal representation in our method. The resulted matrix contains sequential information from different modalities that are represented by low level features with distinct arithmetic properties. For that reason, it is difficult for a learning algorithm to learn multimodal relationships and thus a proper normalization is needed in order to map the different vectors of each modality in an universal arithmetic range. In this work, we have selected to apply a sequence of standard and min-max scaling as a normalization procedure of the multimodal matrix $X$ in order to produce the $X_n$ (ie. normalized $X$) matrix.

\subsection{Representation Learning}
 $X_n$ is a two-dimensional representation of a multimodal sequence. By applying Convolutional Networks with 2-D kernels on $X_n$, we extract local information that is able to capture both uni-modal and cross-modal dynamics across time. More specifically, the kernels of the first layers  mostly model unimodal dynamics that range in neighboring timestamps, while kernels of deeper layers are expected to be able to model cross-modal interactions across a wider range of time. Of course, for each task, there is an optimal balance between short-term unimodal and long-term multimodal modeling that can be found during the training procedure. Based on this formulation of the initial multimodal language sequences, we then select to train unsupervised Convolutional architectures, in order to extract universal representations for this problem. A common choice in the literature is that of Convolutional Autoencoders which have been proven to be effective in image-associated representation learning tasks. More details on the specifics of the Convolutonal AE we trained can be found in the experiments section. 

\subsection{Downstream Classification}

The Encoder part of the trained Convolutional AE can serve as a feature extractor that maps the $N$ x $(M_{audio} + M_{vision} + M_{text})$ multimodal matrix to a $K$ sized feature vector of the AE code. If the learned embeddings have high representational power, the application of a basic Machine Learning algorithm would be effective on downstream tasks of Multimodal Language Analysis. In this work, we focus on testing this hypothesis so our proposed methodology is illustrated in figure \ref{fig:architecture}, where the Convolutional AE is firstly fitted (in an unsupervised manner) on Multimodal Language datasets, while Logistic Regression is then used to train a model that maps the resulted embeddings to the class labels of a particular downstream task. Obviously, our test and validation sets are not be part of the AE's training procedure.
\section{Experiments and Results}

\subsection{Experimental Setup}

\begin{table}[h]
\caption{Comparison of binary accuracy and f1 weighted average metrics against literature models for sentiment classification on MOSEI. With ↑ and ↓ we denote the distance from the minimum and the maximum performance respectively\label{mosei_sota}}
\resizebox{.45\textwidth}{!}{\begin{tabular}{ccc}
{ \textbf{Model (Reference, Evaluation)}}                                   & { \textbf{Acc2}} & { \textbf{F1-weighted}} \\ \hline
{ MV-LSTM (\cite{rajagopalan2016extending}, \cite{zadeh2018multimodal})}    & { 76.4}          & { 76.4}                 \\
{ MFN (\cite{zadeh2018memory}, \cite{zadeh2018multimodal})}                 & { 76}            & { 76}                   \\
{ Graph-MFN (\cite{zadeh2018multimodal}, \cite{tsai2019multimodal})}        & { 76.9}          & { 77.0}                 \\
{ RAVEN (\cite{wang2019words}, \cite{tsai2019multimodal})}                  & { 79.1}          & { 79.5}                 \\
{ MCTN (\cite{pham2019found}, \cite{tsai2019multimodal})}                   & { 79.8}          & { 80.6}                 \\
{ CIA (\cite{chauhan-etal-2019-context}, \cite{chauhan-etal-2019-context})} & { 80.4}          & { 78.2}                 \\
{ MulT (\cite{tsai2019multimodal}, \cite{tsai2019multimodal})}              & { \textbf{82.5}} & { \textbf{82.3}}        \\ \hline
{ Average}                                                                  & { 77.6}          & { 76.9}                 \\ \hline
{ CAE-LR (Ours)}                                                            & { 78 (2↑ 4.5↓)}  & { 76.3 (0.3↑ 4.3↓)}    
\end{tabular}}
\end{table}

\begin{table*}
\caption{Comparison of binary accuracy and f1 weighted average metrics for each emotion against literature models for Emotion Recognition on IEMOCAP. With ↑ and ↓ we denote the distance from the minimum and the maximum performance respectively\label{iemocap_sota}}
\resizebox{\textwidth}{2.5cm}{
\begin{tabular}{ccccccccc}
\multicolumn{1}{l}{{ }}                                            & \multicolumn{2}{c}{{ \textbf{Happy}}}                              & \multicolumn{2}{c}{{ \textbf{Sad}}}                               & \multicolumn{2}{c}{{ \textbf{Angry}}}                             & \multicolumn{2}{c}{{ \textbf{Neutral}}}                              \\ \cline{2-9} 
{ \textbf{Model (Reference, Evaluation)}}                          & { \textbf{Acc2}}     & { \textbf{F1-weighted}} & { \textbf{Acc2}}    & { \textbf{F1-weighted}} & { \textbf{Acc2}}    & { \textbf{F1-weighted}} & { \textbf{Acc2}}       & { \textbf{F1-weighted}} \\
{ DF  (\cite{nojavanasghari2016deep}, \cite{wang2019words})}       & { 86}             & { 81}                & { 81.8}            & { 81.2}                & { 75.8}            & { 65.4}                & { 59.1}               & { 44}                \\
{ MV-LSTM (\cite{rajagopalan2016extending}, \cite{wang2019words})} & { 85.9}             & { 81.3}                & { 80.4}            & { 74}                & { 85.1}            & { 84.3}                & { 67}               & { 66.7}                \\
{ BC-LSTM (\cite{poria2017context}, \cite{wang2019words})}         & { 84.9}             & { 81.7}                & { 83.2}            & { 81.7}                & { 83.5}            & { 84.2}                & { 67.5}               & { 64.1}                \\
{ MARN (\cite{zadeh2018multi}, \cite{wang2019words})}              & { 86.7}             & { 83.6}                & { 82}            & { 81.2}                & { 84.6}            & { 84.2}                & { 66.8}               & { 65.9}                \\
{ TFN (\cite{zadeh2017tensor}, \cite{zadeh2017tensor})}              & { -}             & { 83.6}                & { -}            & {82.8 }                & { -}            & { 84.2}                & {}               & { 65.4}                \\
{ RMFN (\cite{liang2018multimodal}, \cite{wang2019words})}         & { 87.5}             & { 85.8}                & { 83.8}            & { 82.9}                & { 85.1}            & { 84.6}                & { 69.5}               & { 69.1}                \\
{ MFN (\cite{zadeh2018memory}, \cite{wang2019words})}              & { 90.2}             & { 85.8}                & { 88.4}            & { 86.1}                & { \textbf{87.5}}   & { 86.7}                & { 72.1}               & { 68.1}                \\
{ MCTN (\cite{pham2019found}, \cite{tsai2019multimodal})}          & { 84.9}             & { 83.1}                & { 80.5}            & { 79.6}                & { 79.7}            & { 80.4}                & { 62.3}               & { 57}                \\
{ MulT (\cite{tsai2019multimodal}, \cite{tsai2019multimodal})}     & { \textbf{90.7}}    & { \textbf{88.6}}       & { 86.7}            & { 86}                & { 87.4}            & { \textbf{87}}       & { \textbf{72.4}}      & { \textbf{70.7}}       \\
{ ICCN (\cite{sun2020learning}, \cite{sun2020learning})}           & { 87.4}             & { 84.7}                & { \textbf{88.6}}   & { \textbf{88}}       & { 86.3}            & { 85.9}                & { 69.7}               & { 68.5}                \\
\hline
{ Average}                                                         & { 87.13}             & { 83.92}                & { 83.93}            & { 82.35}                & { 83.89}            & { 82.69}                & { 67.38}               & { 63.95}                \\ \hline
{ CAE-LR (Ours)}                                                   & { 85.8  (0.9↑ 4.9↓)} & { 81.60 (0.6↑ 7↓)}      & { 82.3 (3.4↑ 6.3↓)} & { 82.13 (8.13↑ 5.87↓)}  & { 85.6 (9.8↑ 1.9↓)} & { 85.5 (20.1↑ 1.5↓)}    & { 63.85 (4.75↑ 8.55↓)} & { 60.91 (16.91↑ 9.79↓)}
\end{tabular}}
\end{table*}

The MOSEI \cite{zadeh2018multimodal} and IEMOCAP \cite{busso2008iemocap}  datasets have been used for representation learning, as well as Sentiment Analysis and Emotion Recognition respectively. The processed version of IEMOCAP consists of 7318 segments of recorded dyadic dialogues annotated for the presence of the human emotions happiness, sadness, anger and neutral, while MOSEI is a large scale sentiment analysis dataset made up of 22,777 movie review video clips from more than 1000 online Youtube speakers. The data and feature extraction, as well as the train, validation and test splits were obtained from the widely used in the literature CMU-MultimodalSDK repository (\href{https://github.com/A2Zadeh/CMU-MultimodalSDK}{https://github.com/A2Zadeh/CMU-MultimodalSDK}). 

After feature extraction was performed (COVAREP \cite{degottex2014covarep} for audio, GloVe \cite{pennington2014glove} for text and Facet \cite{imotions} for visual) each segment was represented by a 74-d acoustic, a 35-d visual and 300-d textual feature vector. With the use of the aforementioned word-level alignment and concatenation, a matrix, $X$, of 20 x 409 dimensions is obtained for each utterance. After performing a sequence of standard and min-max scaling for each of the 409 features across all 20 timestamps and all dataset instances, the 2-dimensional inputs are properly prepossessed.

In order to train the multimodal representations in an unsupervised manner, we initialized a Convolutional AE with a 4-layer Encoder and its corresponding 4-layer Decoder. The kernel size of the first 2 layers of the Encoder was 3x3, while for the last two layers two 5x5 kernels were used. 2x2 padding along with 1x1 stride and 2x2 max-pooling were used in all layers. The channels of the layers were 32, 64, 128 and 10 respectively. This yields in a (flattened)  representation of 250 dimensions in the code part of the AE. For better representation ability we used the Gelu activation function, while batch normalization was included across all layers. We trained the Convolutional AE using the mean-squared error as loss function, while Adam was chosen to be the optimizer with initial learning rate of 0.002 and a reduce-on-plateau learning rate scheduler. We also performed normal weight initialization and early stopping based on the mean-squared error of the validation set.

For the unsupervised experimental results, we gathered the train sets of MOSEI and IEMOCAP to form the train set and the corresponding validation sets in order to create a multi-dataset validation set. For different kinds of pre-training as well as the performance of the Convolutional AE in terms of the mean-squared error function, we refer the reader to section \ref{ablation_study}.

\subsection{Results on Downstream Classification}

Using the Encoder part of the Convolutional AE as a feature extractor, we get a 250-d feature vector for each utterance. This means that the initial 20 x 409 (=8180 feature values) matrix is reduced to a 250-d vector (32 times less elements). These embeddings have been used to train a Logistic Regression model for the two downstream tasks of Sentiment Analysis and Emotion Recognition. Following the existing literature \cite{zadeh2018multi, zadeh2018memory, zadeh2017tensor, wang2019words, zadeh2018multimodal, liu2018efficient, tsai2019multimodal, tsai2019learning,  chauhan-etal-2019-context, sun2020learning}, we report the binary accuracy and weighted averaged f1 metrics on sentiment for MOSEI, and on each of the 4 emotions of IEMOCAP in an one-vs-all manner.

In tables \ref{mosei_sota} and \ref{iemocap_sota} we compare our results with state-of-the-art and well established architectures in the tasks of Sentiment Analysis and Emotion Recognition. More specifically, we compare to MV-LSTM \cite{rajagopalan2016extending}, MFN \cite{zadeh2018memory}, Graph-MFN \cite{zadeh2018multimodal}, RAVEN \cite{wang2019words}, MCTN \cite{pham2019found}, CIA \cite{chauhan-etal-2019-context}, MulT \cite{tsai2019multimodal}, DF \cite{nojavanasghari2016deep}, BC-LSTM \cite{poria2017context}, MARN \cite{zadeh2018multi}, TFN \cite{zadeh2017tensor}, RMFN \cite{liang2018multimodal} and ICCN \cite{sun2020learning} which, with the use of the CMU-MultimodalSDK repository, are trained and evaluated on the exact same sets. For each model we reference both the original work and the one that evaluated the method on the MOSEI or IEMOCAP datasets. For detailed explanation and comparison of the aforementioned architectures, we refer the reader to detailed reviews on Multimodal Sentiment Analysis \cite{gkoumas2021makes} and Multimodal Emotion Recognition \cite{app11177962}. 

As seen in \ref{mosei_sota} and \ref{iemocap_sota} the proposed generic mutlimodal representations achieve competitive performance with the use of a basic Machine Learning algorithm (Logistic Regression). More specifically, our lightweight method achieves an average (calculated on state-of-the art competitive models) performance on sentiment analysis with 2 absolute points above minimum and 4.5 below maximum binary accuracy. It also performs close to average and always above minimum for the task of Emotion Recognition. Therefore, a very simple simple classification algorithm can score the average performance of the SotA models, which clearly indicates that the learned multimodal language embeddings have strong representation power and can be used across different multimodal language tasks, despite the different recording set-up across datasets. The code for all experiments can be found in the mlr repository (\href{https://github.com/lobracost/mlr}{https://github.com/lobracost/mlr}). 

\subsection{Ablation Study}\label{ablation_study}

\begin{table}
\centering
\caption{Convolutional AE pretraining for different modalities and dataset combinations\label{cae_pretrain}}
\resizebox{.45\textwidth}{!}{
\begin{tabular}{ccc}
{ \textbf{Modalities}}   & { \textbf{Embeddings Training}} & { \textbf{MSE} ( x $10^{-4}$)}  \\ \hline
{ Audio}                 & { MOSEI \& IEMOCAP}  & { 32.34} \\
{ Vision}                & { MOSEI \& IEMOCAP}  & { 22.67} \\
{ Text}                  & { MOSEI \& IEMOCAP}  & { 44.22} \\
{ [Audio, Vision]}        & { MOSEI \& IEMOCAP}  & { 30.21}  \\
{ [Vision, Text]}         & { MOSEI \& IEMOCAP}  & { 24.34} \\
{ [Audio, Text]}          & { MOSEI \& IEMOCAP}  & { 29.94} \\
{ [Audio, Vision, Text]} & { MOSEI \& IEMOCAP}  & { 24.11} \\
{ [Audio, Vision, Text]} & { MOSEI}            & { 23.12} \\
{ [Audio, Vision, Text]} & { IEMOCAP}              & { 20.58}
\end{tabular}}

\centering
\caption{Performance on MOSEI sentiment classification using embeddings trained on different modality combinations\label{modalities_pretrain}}
\begin{tabular}{ccc}
{ \textbf{Modalities}}   & { \textbf{Acc2}}  & { \textbf{F1-weighted}} \\ \hline
{ Audio}                 & {71.06}          & {59.1}                \\
{ Visual}                & {71.06}          & {59.1}                \\
{ Text}                  & {75.4}          & {72.63}                \\
{ [Audio, Visual]}       & {71.06}          & {59.87}                \\
{ [Visual, Text]}        & {71.04}          & {59.01}                \\
{ [Audio, Text]}         & {77.86}          & {76.09}                \\
{ [Audio, Visual, Text]} & { \textbf{78}} & { \textbf{76.3}}      
\end{tabular}

\centering
\caption{Performance on MOSEI sentiment classification using embeddings trained on different dataset combinations\label{dataset_pretrain}}
\begin{tabular}{ccc}
\multicolumn{1}{l}{{ \textbf{Pretrained representations}}} & { \textbf{Acc2}}  & { \textbf{F1-weighted}} \\ \hline
{ MOSEI}                                                   & { 77.2}          & { 75.4}                \\
{ IEMOCAP}                                                 & { 76.7}          & { 74.5}                \\
{ MOSEI \& IEMOCAP}                                        & { \textbf{78}} & { \textbf{76.3}}      
\end{tabular}

\end{table}

In this section, we examine the role of the modalities combinations, as well as different datasets used for the unsupervised training of the embeddings. To this end, we trained the Convolutional AE on a range of different modality  and training data combinations. For each combination, we report the mean-squared error in table \ref{cae_pretrain}. In that way we can gain an insight on the quality of the learnt information compression for each modality combination. It is easily derived that the learnt embeddings of our method are more informative with respect to the original visual modality and less collective for the textual one.

\subsubsection{The role of different modalities}

Using the produced embeddings of each modality combination in order to perform downstream classification, we end up with the results of table \ref{modalities_pretrain} for the MOSEI dataset. As it can be clearly seen \textit{(i)} using all three modalities enriches the representation power, and \textit{(ii)} the textual modality is the most informative for the task of Multimodal Language Analysis, a fact that is also known in the literature \cite{sun2020learning}. Thus our method achieves to effectively express unimodal information and learn multimodal interactions of temporal language sequences. 

\subsubsection{Cross-domain generalization ability}\label{cross-dataset}

In order to examine whether information of one dataset can be used to other tasks, we performed different pretraining to the Convolutional Autoencoder and record the performance on downstream classification for the MOSEI dataset. The two crucial remarks from table \ref{dataset_pretrain} that make our work widely useful is that \textit{(i)} our methodology leads to embeddings that can easily generalize to new data that have been recorded in a different way and for different tasks, since the performance for Sentiment Analysis when using embeddings pretrained on IEMOCAP is 0.5 absolute points (in terms of binary accuracy) below the ones trained in the MOSEI train data and \textit{(ii)} our embeddings can be enriched from information of different datasets, since the classification using the embeddings trained on MOSEI \& IEMOCAP performs better than the ones trained on just the MOSEI training set. That clearly indicates that the proposed Encoder can serve as feature extractor in a range of Multimodal Language tasks and used for generalization in unseen data formats.

\subsection{Model Complexity}

\begin{table}
\centering
\caption{Comparison of model parameters\label{complexity_comparison}}
\begin{tabular}{ccc}
{ \textbf{}}                          & \multicolumn{2}{c}{{ \textbf{Number of Parameters}}}        \\
{ \textbf{Model}}                     & { \textbf{MOSEI}} & { \textbf{IEMOCAP}} \\ \hline
{ TFN}                                & { 6,804,859}      & { 23,198,398}       \\
{ RMFN}                               & { -}              & { 1,732,884}        \\
{ MFN}                                & { 415,521}        & { 1,325,508}        \\
{ MulT}                               & { 874,651}        & { 1,074,998}        \\
{ CAE-LR (ours)} & { 256,453}        & { 257,206}         
\end{tabular}
\end{table}

In \cite{gkoumas2021makes} the authors retrained 11 of the most powerful and widely used models for Multimodal Language Analysis and list the number of parameters for some of them. However, this study, due to different pretraining, reports smaller amount of parameters for some models (eg. \cite{mai2021analyzing} reports 1,549,321 parameters for MulT on the MOSEI dataset with a different pretraining procedure). However, for the sake of a complete comparison, we choose to report the results of \cite{gkoumas2021makes}, though this may underestimate the number of parameters of the reported models. In table \ref{complexity_comparison} we report the number of parameters of the proposed methods against the models reported in \cite{gkoumas2021makes}. It can be noticed that the Encoder used for feature extraction uses the same number of parameters across both datasets, which is not the case for other architectures where the parameter difference across tasks ranges from ~123\% to ~340\% (probably due to the number of classes). That is a direct outcome from the fact that our method does not require extensive fine-tuning techniques in order to be generalized to other tasks. Combining the feature extraction (256,202 parameters for both tasks) with the inference part (251 parameters for MOSEI and 1004 for IEMOCAP) we end up with an end-to-end method that includes parameters in the range 256,202-257,206 while the rest architectures are highly greedy in terms of parameter amounts since they range from 415,521 to 23,198,398. Thus the CAE-LR is a lightweight method that can be easily generalized to other tasks, without impact in the number of parameters (shown in Table \ref{complexity_comparison}), nor significant performance loss (shown in section \ref{cross-dataset}).
\section{Conclusion}

In this work, we have presented a method for extracting unsupervised Multimodal Language representations using 2-D aligned multimodal sequences and Convolutional Autoencoders, in a totally unsupervised learning process. Extensive experimentation in Sentiment Analysis and Emotion Recognition prove that the performance of the proposed method is competitive related to the SotA, though the respective representation model is much smaller (around 200K parameters). Apart from being extremely lightweight, the proposed architecture is transferable to other domains without negative impact on the performance: experiments prove that transferring knowledge from one dataset to another, without retraining -not even tuning- the representation model itself (just training a logistic regression classifier on the extracted embeddings), does not significantly affect the classification performance in the target domain. Finally, the proposed representation method is openly available through an easy-to-use model (\href{https://github.com/lobracost/mlr}{https://github.com/lobracost/mlr}) which can be directly applied to any similar Multimodal Language Modeling downstream task. The proposed approach could be further enriched in a future work, by training multimodal embeddings on a greater amount of datasets and reporting performance on more Multimodal Language tasks, such as Persuasiveness Prediction.


\comment{
\begin{figure}[htb]

\begin{minipage}[b]{1.0\linewidth}
  \centering
  \centerline{\includegraphics[width=8.5cm]{image1}}
  \centerline{(a) Result 1}\medskip
\end{minipage}
\begin{minipage}[b]{.48\linewidth}
  \centering
  \centerline{\includegraphics[width=4.0cm]{image3}}
  \centerline{(b) Results 3}\medskip
\end{minipage}
\hfill
\begin{minipage}[b]{0.48\linewidth}
  \centering
  \centerline{\includegraphics[width=4.0cm]{image4}}
  \centerline{(c) Result 4}\medskip
\end{minipage}
\caption{Example of placing a figure with experimental results.}
\label{fig:res}
\end{figure}
}



\bibliographystyle{IEEEbib}
\bibliography{main}

\end{document}